\pdfoutput=1
\documentclass[11pt]{article}

\usepackage[review]{acl}

\usepackage{times}
\usepackage{latexsym}
\usepackage[review]{acl}
\usepackage{times}
\usepackage{latexsym}
\usepackage{tipa}
\usepackage{amsmath}
\usepackage{amsfonts} 
\usepackage{booktabs}
\usepackage{multirow}
\usepackage{adjustbox}
\usepackage{longtable}
\usepackage{graphicx}
\usepackage{float} 
\usepackage[T1]{fontenc}
\usepackage[utf8]{inputenc}

\usepackage{microtype}

\usepackage{inconsolata}

\usepackage{adjustbox}
\usepackage{afterpage}

\title{Learning Beyond Limits:  Multitask Learning and Synthetic Data for Low-Resource Canonical Morpheme Segmentation}
\usepackage{lineno}
\nolinenumbers
\author{
Changbing Yang, Garrett Nicolai\\
University of British Columbia \\
{\tt cyang33@mail.ubc.ca }
}

\begin{document}
\maketitle
\begin{abstract}

We introduce a transformer-based morpheme segmentation system that augments a low-resource training signal through multitask learning and LLM-generated synthetic data. Our framework jointly predicts morphological segments and glosses from orthographic input, leveraging shared linguistic representations obtained through a common documentary process to enhance model generalization. To further address data scarcity, we integrate synthetic training data generated by large language models (LLMs) using in-context learning. Experimental results on the SIGMORPHON 2023 dataset show that our approach significantly improves word-level segmentation accuracy and morpheme-level F1-score across multiple low-resource languages.

\end{abstract}

\section{Introduction}

Morphological segmentation—the process of breaking words into their smallest meaningful units—is a fundamental task in linguistic analysis.  This process has two goals: first, to identify morpheme boundaries, and second, to restore phonological changes between canonical and surface forms.
%and stripping away inflectional or derivational modifications. 
For example, the word \textit{happiness} is composed of two surface morphemes:  \textit{happi} + \textit{-ness}. Underlyingly, the root \textit{happy} undergoes an orthographic modification when it combines with \textit{-ness}.  Canonical segmentation produces the normalized \textit{happy-ness}.  

%\begin{center}
%\begin{adjustbox}{width=0.5\linewidth}
%    \begin{tabular}{ll}
%        \textbf{Word:} & happiness \\
%        \textbf{Surface:} & happ\color{red}{i} -ness\\
%        \textbf{Canonical:} & happ\color{green}y -ness \\
%    \end{tabular}
%\end{adjustbox}
%\end{center}

Canonical segmentation is particularly critical for analyzing low-resource and morphologically-complex languages. Linguistic documentation relies on language experts creating Interlinear Glossed Texts (IGT). 
%a standardized framework used by linguists for language documentation. 
An IGT entry consists of four tiers: 1. orthographic text, the original sentence; 2. morpheme segmentation, decomposing words into canonical morphemes; 3. glossing, assigning linguistic labels to each morpheme; and 4. translation, providing an equivalent sentence in a high-resource matrix language like English. An example from Gitksan follows.
\begin{center}
\begin{adjustbox}{width=\columnwidth}
    \begin{tabular}{ll}
        \textbf{Orthography:} & Ii hahla'ls\textcolor{red}{di}'y goohl IBM \\
        \textbf{Segmentation:} & ii hahla'ls\textcolor{green}{t}-'y goo-hl IBM \\
        \textbf{Gloss:} & CCNJ work-1SG.II LOC-CN IBM \\
        \textbf{Translation:} & And I worked for IBM.\\
    \end{tabular}
\end{adjustbox}
\end{center}

The construction of IGTs is a process that requires significant linguistic expertise. For languages with few speakers, the segmentation step alone can be a complex and time-consuming task.
%As the fundamental step for further study of language structure and meaning, manual segmentation of each word in the Interlinear Glossed Text (IGT) is a labor-intensive process. 
Previous research has begun to automate this process using neural models \cite{kann-etal-2016-neural, ruzsics-samardzic-2017-neural, wang2019neural, rice-etal-2024-tams}, but performance remains limited by scarce annotated training data. Most approaches focus exclusively on segmenting the orthographic tier \cite{kann-etal-2016-neural, ruzsics-samardzic-2017-neural, wang2019neural}. \newcite{rice-etal-2024-tams}, however explore augmenting the segmentation signal with an additional encoder tied to the translation tier. This method depends on manual word alignment between source and translated text, and does not ease the need for linguistic expertise. We instead propose two methods for leveraging existing signals to improve canonical segmentation in low-resource language documentation:

%We approach the problem from a different direction: instead of using another tier as input to the segmentation task, we instead add the glossing tier as output of a multi-task objective. Glossing inherently contains rich morphological and semantic information, often encapsulating much of the detail found in translations. In this paper, we propose two methods for improved canonical segmentation in low-resource language documentation:

\textbf{Multitask learning} 
%A multi-task learning framework that integrates canonical segmentation and glossing within a single model. 
Multitask learning encourages generalization across complementary objectives \cite{caruana1997multitask}, and can enhance robustness in low-resource scenarios \cite{lin-etal-2018-multi-lingual, johnson-etal-2017-googles}. In our framework, the model is trained to jointly predict the segmentation and glossing tiers of an IGT, with only the orthographic tier as input. Incorporating glossing as a parallel objective in multitask learning can exploit beneficial information without necessitating further data curation, as glossing is already a component of IGT. By learning these related tasks simultaneously, the model gains access to rich linguistic information —morpheme boundaries from the segmentation tier, and labels from the glossing tier.
%effectively amplifying the utility of limited training data. 
%This shared representation enables more robust training than a single objective.
%the model to infer segmentation patterns indirectly from glossing annotations, even when explicit segmentation examples are sparse.
%Moreover, unlike previous methods that rely on manual matching of word alignments, our approach eliminates the need for such labor-intensive steps.

\textbf{LLM synthetic data}
The scarcity of annotated datasets for low-resource languages often causes neural models to overfit frequent character sequences rather than generalizing to true morphological structures, a phenomenon known as label bias \cite{wiseman2016sequence}. To address this, we supplement the training data with synthetic examples created by large language models (LLMs) with in-context learning. Since canonical segmentation involves resolving phonological alternations (e.g., mapping hahla'lsdi to -hahla'lst-), LLMs excel at this task by learning and replicating these alternations directly from interlinear glossed text (IGT) examples—without requiring explicit rule encoding. 
%LLMs learn to replicate these transformations from interlinear glossed text (IGT) examples while preserving morphological integrity. 
By systematically varying the proportion of synthetic data, we assess its role in mitigating data scarcity while maintaining segmentation consistency.

Our contributions are as follows:
\begin{itemize}
\item We introduce a multitask learning framework that jointly learns to segment and gloss, improving segmentation performance across multiple low-resource languages.
\item We synthesize data to augment sparse training data for segmentation and evaluate its effectiveness at different saturation levels.
\item We combine the two strategies, demonstrating that multitask learning and synthetic data complement each other to enhance segmentation quality.
\end{itemize}

%Our contributions are threefold: (1) We propose a multi-task learning framework that jointly predicts segmentation and glossing, improving segmentation performance across multiple low-resource languages. (2) We introduce synthetic data generation to supplement sparse training data and evaluate its effectiveness at varying saturation levels. (3) We compare different training strategies, showing that multi-task learning and synthetic data complement each other to enhance segmentation quality.

\section{Experiment Setup and Methodology}

Following the work of \cite{rice-etal-2024-tams}, we conduct experiments in the languages of the SIGMORPHON 2023 Shared Task dataset \cite{ginn2023findings}\footnote{The dataset is licensed under CC BY-NC 4.0.}. The TAMS system proposed by \newcite{rice-etal-2024-tams} requires a manual alignment between source and matrix language, and therefore, linguistic expertise, limiting their results to a subset of the dataset's languages (Arapaho, Lezgi, and Tsez), for which we use the same data splits. We expand our experiments to the remaining languages in the data, including Gitksan, Nat\"ugu, Nyangbo, and Uspanteko. Data is split by identifying all unique words in each language dataset, and re-split using the same 6:2:2 split in the TAMS paper\footnote{Our splits will be made available after publication.}.  Specifics for each language are in Table \ref{tab:lang}.

\subsection{Multitask Model for Canonical Segmentation}
We treat canonical segmentation as a sequence-to-sequence task and conduct our experiments with a modified version of Fairseq’s \cite{ott-etal-2019-fairseq} implementation of transformers \cite{vaswani2017attention}. We modify the transformer architecture with a multitask objective \footnote{Our code is available: \url{https://link/to/our/repo}. Our implementation is modified based on \newcite{zhou-etal-2019-improving}'s work: \url{https://github.com/shuyanzhou/multi-task_transformer}.} . %Unlike the sequential training approach used in (CITATION), our model introduces two parallel decoders that simultaneously rather than training these tasks in separate phases. 
Our model consists of a shared encoder that processes the input word from the orthographic tier, generating a latent representation. This representation then serves as input to a pair of decoders: the first learns to produce a canonical segmentation and the other generates the corresponding gloss\footnote{If the word gloss is "work-1SG.II", the gloss decoder will generate it as "w-o-r-k-1SG.II"}. We define a joint loss function as the weighted sum of segmentation loss and glossing loss:

\begin{equation}
\mathcal{L}_{total} = \lambda \mathcal{L}_{seg} + (1 - \lambda) \mathcal{L}_{gloss}
\end{equation}

where the segmentation loss weight \( \lambda \) is tuned within the range of 0.8 to 1\footnote{Appendix \ref{weighting} illustrates the impact of \( \lambda \) on Lezgi model performance.}, while the weight of the glossing objective is complemetary, ensuring that the model prioritizes segmentation accuracy while still leveraging glossing information as auxiliary supervision. %This balance allows the model to optimize for segmentation while benefiting from additional supervision from gloss annotations. 
Hyper-parameters and model details are in the Appendix \ref{model-appendix}.

\begin{table}
\begin{adjustbox}{width=\columnwidth,center}
\begin{tabular}{lllll}
\toprule
Language & Train & Dev & Test & Matrix lang.  \\\midrule
Arapaho (arp) & 16666 & 10760 & 9849 & (eng) \\
Gitksan (git) & 323 & 107 & 109 & (eng) \\
Lezgi (lez) & 1236 & 412 & 412 & (eng) \\
Nat\"ugu (ntu) & 1953 & 651 & 652 & (eng) \\
Tsez (ddo) & 3,558 & 445 & 445 & (eng) \\
Uspanteko (usp) & 7033 & 2345 & 2344 & (spa) \\
Nyangbo (nyb) &1499 & 499& 501 &- \\
\bottomrule
\end{tabular}
\end{adjustbox}
\caption{2023 SIGMORPHON Shared Task Dataset \cite{ginn2023findings}}
\label{tab:lang}
\end{table}

\subsection{Generating Synthetic Examples}
To address data scarcity, we generate synthetic segmentation data using GPT-4o with in-context learning to supplement the limited training data.
%Our goal is to expose the model to various morphophonological alternations, which are essential for accurate canonical segmentation in morphologically complex languages.

First, we extract all words from the training data which have a disjunction between their underlying and surface morphemes. 
%This may involve vowel alternations, consonant mutations, or other phonological modifications.
These forms will serve as in-context examples for the LLM.%These words serve as the foundation for generating synthetic examples that emphasize morphophonological transformations.

%Next, we identify words in the training set that share a stem with our learning examples. We pair these words into a triple: (stem, word, segmentation), and construct a prompt that demonstrates how canonical segmentation alters word forms while preserving their morphological structure. The prompt example of Nat\"ugu is shown in Appendix.

Next, we construct a structured prompt that includes: 1. A word stem and its meaning. 2. Example words from the training data that share this stem, along with their canonical segmentations and glosses. 3. A list of grammatical morphemes\footnote{Grammatical morphemes are functional elements in language that indicate grammatical relationships such as tense, number, case, or person, rather than carrying lexical meaning, as seen in markers like 1SG.II (first-person singular) and LOC (locative) in the IGT example.} and their corresponding glosses, extracted directly from the training data. 
%By including glossing information, we ensure that the generated examples adhere to linguistic conventions and maintain morphological consistency.

The LLM then generates new words by combining the stem with grammatical morphemes, applying morphophonological alternations based on the examples provided.The resulting triples—surface form, canonical segmentation, and gloss—approximate IGT text and expand the model’s morphological coverage. An example prompt for Nat\"ugu is in Appendix~\ref{LLM-prompt-example}.

%Finally, the LLM generates synthetic word forms based on the provided examples. These newly created samples are then incorporated into the training dataset, expanding the available segmentation pairs without requiring additional manual annotations. 

\section{Results and Findings}

We now discuss the findings of our experiments. Following TAMS \cite{rice-etal-2024-tams}, we evaluate across 3 metrics: word-level accuracy, morpheme-level F1, and the sum of edit-distances across all test instances.  We evaluate against reported results from the TAMS paper, as well as a Fairseq baseline with a single decoder devoted to segmentation.

\subsection{Multitask Learning Performance}

%We first evaluate our multi-task setup on each language described previously. We evaluate the same metrics as TAMS: word-level accuracy, morpheme-level F1 score, and the sum of edit distance. Along with the results reported in the TAMS paper (two methods), we also evaluate against a single-task baseline, also implemented in Fairseq.
\begin{table}[h]
\begin{adjustbox}{width=\columnwidth,center}
    \centering
    \begin{tabular}{|l|l|c|c|c|c|c|c|c|c|}
        \hline
        \textbf{Model} & \textbf{Metric} & \textbf{lez} & \textbf{ddo} & \textbf{arp} & \textbf{git} & \textbf{ntu} & \textbf{nyb} & \textbf{usp} & \textbf{ave} \\
        \hline
        \multirow{3}{*}{Baseline}  
        & ACC↑ & 44.66 & \textbf{82.6} & 67.08 & 47.71 & 63.04 & \textbf{80.48} & 55.05 & 62.95 \\  
        & F1↑  & 60.75 & 90.44 & 81.11 & 65.5  & 80.3  & 90.24 & 75.66 & 77.71 \\  
        & ED↓  & 568   & 652   & 10495 & 117   & 458   & 154   & 1799 & 2034.71 \\  
        \hline
        \multirow{3}{*}{\shortstack{TAMS }}

        & ACC↑ & 46.84 & 80.78 & 67.72 & -- & -- & -- & -- & -- \\  
        & F1↑  & 62.48 & 89.52 & 81.62 & -- & -- & -- & -- & -- \\  
        & ED↓  & 532   & 701   & 9899  & -- & -- & -- & -- & -- \\  
        \hline
       \multirow{3}{*}{\shortstack{TAMS-CLS }}
        & ACC↑ & 47.09 & 81.96 & 67.4  & -- & -- & -- & -- & --\\  
        & F1↑  & 62.48 & 90.08 & 81.45 & -- & -- & -- & -- & --\\  
        & ED↓  & 537   & \textbf{643}   & 9970  & -- & -- & -- & -- & --\\  
        \hline
        \multirow{3}{*}{Multitask }  
        & ACC↑ & \textbf{48.54} & 82.51 & \textbf{78.01} & \textbf{52.29} & \textbf{68.87} & 79.84 & \textbf{56.12} & \textbf{66.59} \\  
        & F1↑  & \textbf{68.84} & \textbf{92.12} & \textbf{84.14} & \textbf{71.64} & \textbf{84.09} & \textbf{91.43} & \textbf{77.18} & \textbf{81.35} \\  
        & ED↓  & \textbf{519}   & 698   & \textbf{6543}  & \textbf{112}   & \textbf{373}   & \textbf{149}   & \textbf{1623} & \textbf{1431} \\  
        \hline 
        \hline
    \end{tabular}
    \end{adjustbox}
    \caption{Comparison of canonical segmentation models across multiple languages. Each model includes three sub-rows for ACC, F1, and ED, with the last column showing average metrics. \textbf{Bolded} values indicate language bests for each metric. ↓ indicates that lower is better.}
    \label{tab:MT_results}
\end{table}

Table \ref{tab:MT_results} demonstrates that the multitask model achieves superior overall performance.  Most languages see improvements over the best alternative.  Furthermore, attaching a multitask objective improves over the single-task objective for each metric, on average.  Languages which already have higher performance, such as Nyangbo and Tsez, still see improvements at the morpheme level, although Nyangbo demonstrates that improvements in F1 are not always accompanied by a similar improvement in accuracy.
It is possible that the benefits of multitask learning may be more significant at the morpheme level than at the word level. 
%This may be due to the nature of Nyangbo’s morphology, where segmenting words into morphemes is more challenging than recognizing words themselves. Similarly, Uspanteko demonstrates modest but consistently better segmentation outputs.

Training data size seems to have little impact on the benefits of multitask learning.
Languages such as Arapaho, with significantly more data than the sparsest languages, observes large improvements, while %attaining the highest average word-level accuracy (66.74\% vs. a baseline of 62.37\%), morpheme F1-score (81.35\% vs. 77.86\%), and lowest edit distance (1445.29 vs. 2280.43). While the baseline single-task approach performs strongly on individual languages like Tsez (82.6\% ACC) and Nyangbo (80.48\% ACC), the multi-task framework shows broader cross-linguistic robustness, achieving state-of-the-art results in 5/7 languages for ACC and 6/7 for F1. Notably, it substantially improves morphological segmentation for languages like Arapaho (78.01\% ACC vs. baseline 67.08\%) while maintaining strong performance on analytic languages like Nat\"ugu (68.87\% ACC). Although the TAMS variants show modest improvements over the baseline, our multi-task approach provides further gains along every metric. Edit distance patterns correlate with accuracy metrics, with the multi-task model reducing morphological segmentation errors by 36.6\% on average compared to the baseline, demonstrating its effectiveness at learning canonical changes.
%Beyond the three languages evaluated by TAMS, we extend our experiments to Gitksan, Nat\"ugu, Nyangbo, and Uspanteko to assess the generalizability of multi-task learning. The results indicate that the second task provides benefits in most of these languages, particularly in segment accuracy and morpheme precision. 
Gitksan and Nat\"ugu, which have much less training data, also improve when a multitask objective is introduced.

%Our initial focus is on three languages—Lezgi, Tsez, and Arapaho—previously studied in the TAMS paper, followed by an extended evaluation on four additional languages: Gitksan, Nat\"ugu, Nyangbo, and Uspanteko. For the three benchmark languages, multi-task learning consistently improves performance across key metrics. In Lezgi and Arapaho, multi-task learning leads to noticeable gains in both word-level accuracy and morpheme segmentation quality. Particularly in Arapaho, where the language exhibits high morphological complexity, multi-task learning helps the model better capture the underlying morphemic structure. Tsez, which already has high segmentation performance, also benefits from multi-task learning, especially in morpheme segmentation accuracy, although the improvement is more modest compared to the other two languages.

%The observed improvements suggest that including glossing information as an auxiliary supervision signal enhances the model's ability to distinguish morpheme boundaries (as evidenced by the 16$\%$ error reduction in average F1). %This effect is particularly important in morphologically complex languages, where surface segmentation alone may not fully capture underlying morphemes.

A qualitative analysis suggests that multitask learning improves the overall accuracy of morpheme segmentation by reducing unnecessary modifications.  That is, the baseline model is too aggressive in employing textual normalization, making changes where they are not appropriate. In languages with numerous morphological alternations, such as Arapaho and Lezgi, multitask learning significantly reduces edit distances by removing alternations that the baseline deems necessary.
%indicating that the model is producing segmentations that are closer to the gold standard. 
In contrast, in languages with already high segmentation accuracy, such as Tsez, decreases in edit distance are less pronounced - the glossing information may not add much extra signal.

Overall, these findings indicate that integrating glossing information as an extra predictive task improves model quality, without the need for extra annotation. The improvements are particularly noticeable in languages with complex segmentation patterns, demonstrating the effectiveness of this approach in improving canonical segmentation in low-resource settings.

\subsection{Learning Curve of multitask Learning}

After observing in our previous experiments that data size had less of an impact than linguistic constraints, we conducted experiments aimed at further investigating the role that data size plays on multitask learning.
For each language, we create artificially small training sets by limiting the data to 25, 50, 75, and 100$\%$ of the original training set.
The comparison of the average learning curves is presented in Figure \ref{MT_LC}.\footnote{For individual language curves, please see Appendix \ref{learning-curve-all}.} 

\begin{figure}
\includegraphics[width=\columnwidth]{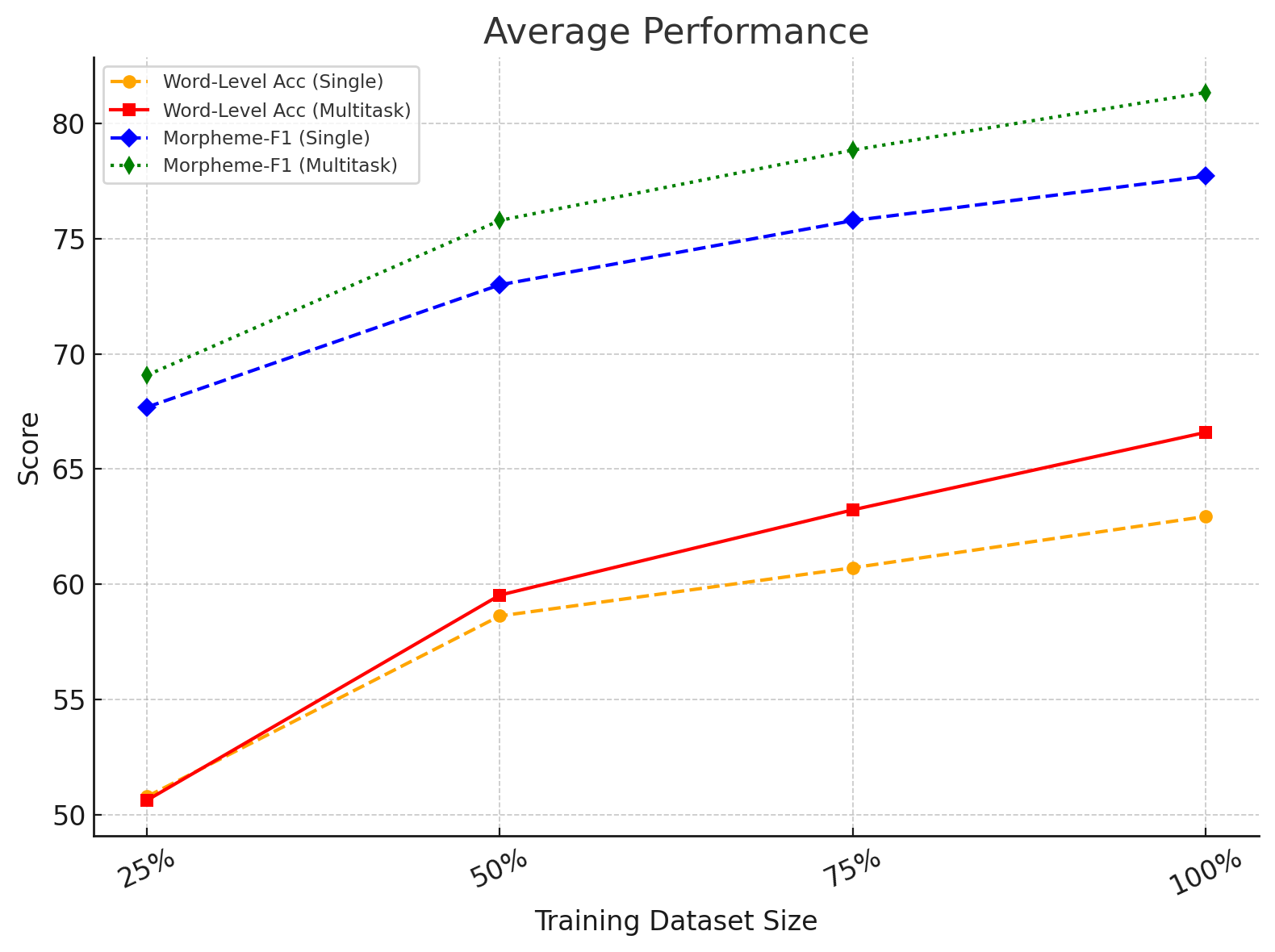}
\caption{The average learning curves for the F1 (top) and Accuracy (bottom) metrics.}
\label{MT_LC}
\end{figure}

We observe that in general, the improvements obtained from multitask learning increase as more training data is available, although there is still an observed benefit in extremely low-data settings.  This is promising, as it suggests that improvements obtained in aiding the documentary process at the beginning will eventually feed a virtuous cycle, with increasing gains as further data is created.

\subsection{Addressing Data Scarcity with LLM-Generated Data}

After observing that the model benefits from extra training data, we seek to augment the training data with synthetic examples.  In our final experiment, we supplement our multitask model with training examples generated by an LLM.  We control the percentage of added synthetic examples - increasing in increments of 25$\%$ of the gold training data.  We report the results in Table \ref{tab:LLM_results}.
%While multi-task learning improves segmentation performance, the learning curve suggests that more data is beneficial. For low-resource languages, however, large annotated datasets are often not available. To supplement the training set, we turn to large language models (LLMs), using in-context learning to provide auxiliary examples.

We observe continued, if modest, improvements when supplementing multitask learning with synthetic data.  Some languages, like Gitksan, only start to improve when the percentage of synthetic examples approaches the number of natural ones.  Other languages, like Arapaho, which already contains much larger data stores, see regular improvements as more data is added.  There do seem to be some limitations to the idea that more data is always better, however; Lezgi sees an improvement only with moderate levels of extra data, and high-performing languages like Tsez and Nyangbo are difficult to improveme any further.  On average, we see similar trends to multitask learning on its own - with most of the benefit coming at the morpheme level.

\begin{table}[h]
\begin{adjustbox}{width=\columnwidth,center}
    \centering
    \begin{tabular}{|l|l|c|c|c|c|c|c|c|c|}
        \hline
        \textbf{Model} & \textbf{Metric} & \textbf{lez} & \textbf{ddo} & \textbf{arp} & \textbf{git} & \textbf{ntu} & \textbf{nyb} & \textbf{usp} & \textbf{ave} \\
        \hline
      % \multirow{3}{*}{Baseline (Singletask)}  
      %  & ACC↑ & 44.66 & \textbf{82.6} & 67.08 & 47.71 & 63.04 & 80.48 & 55.05 & 62.37 \\  
      % & F1↑  & 60.75 & 90.44 & 81.11 & 65.5  & 80.3  & 90.24 & 75.66 & 77.86 \\  
      %  & ED↓  & 568   & \textbf{652}   & 10495 & 117   & 458   & 154   & 1799 & 2280.43 \\  
      %  \hline
     %   \multirow{3}{*}{TAMS}  
    %    & ACC↑ & 46.84 & 80.78 & 67.72 & -- & -- & -- & -- & -- \\  
    %    & F1↑  & 62.48 & 89.52 & 81.62 & -- & -- & -- & -- & -- \\  
    %    & ED↓  & 532   & 701   & 9899  & -- & -- & -- & -- & -- \\  
     %   \hline
    %    \multirow{3}{*}{TAMS-CLS}  
    %    & ACC↑ & 47.09 & 81.96 & 67.4  & -- & -- & -- & -- & --\\  
    %    & F1↑  & 62.48 & 90.08 & 81.48 & -- & -- & -- & -- & --\\  
    %    & ED↓  & 537   & 648   & 9970  & -- & -- & -- & -- & --\\  
     %   \hline
        \multirow{3}{*}{M}  
        & ACC↑ & 48.54 & \textbf{82.51} & 78.01 & 52.29 & 68.87 & 79.84 & 56.12 & 66.59 \\  
        & F1↑  & 68.84 & \textbf{92.12} & 84.14 & 71.64 & 84.09 & \textbf{91.43} & 77.18 & 81.35 \\  
        & ED↓  & 519   & 698   & 6543  & 112   & 373   & 149   & 1623 & 1431 \\  
        \hline 
        %\multirow{3}{*}{Only CRF}  
       % & ACC↑ & 49.76 & 80.0  & 71.1  & 53.21 & 70.25 & 80.84 & 57.61 & 66.11 \\  
        %& F1↑  & 67.04 & 90.96 & 76.91 & 68.8  & 83.48 & 90.77 & 78.47 & 79.20 \\  
       % & ED↓  & 521   & 783   & 9401  & 96    & 328   & 142   & 1492 & 1623.29 \\  
        %\hline
      %  \multirow{3}{*}{multi-task (0.8, 0.2) + CRF}  
      %  & ACC↑ & 47.33 & 80.32 & 77.0  & 50.46 & 67.33 & 78.24 & 56.2 & 65.99 \\  
       % & F1↑  & 67.29 & 91.34 & 83.68 & 68.21 & 82.64 & 90.17 & 78.06 & 80.48 \\  
       % & ED↓  & 556   & 772   & 6844  & 126   & 398   & 163   & 1604 & 1623.29 \\  
      %  \hline
         %\multirow{3}{*}{Singletask + LLM (25\% traindata size)}  
       % & ACC↑ & 48.79 & 80.28 & 78.17 & 53.96 & 66.10&  80.04 & 56.50 & 66.41\\  
       % & F1↑  & 68.17 & 90.82& 83.96 & 70.95 & 80.55 &  90.73 & 76.59 & 80.97\\  
      %  & ED↓  &475   & 852  & 6534 & \textbf{92}    & 357   &  137 & 1544& 1413.00\\  
       % \hline
        
        \multirow{3}{*}{M+ LLM (0.25)}  
        & ACC↑ & 49.27 & 80.41 & 78.14  & 52.29 & \textbf{69.02} &  80.21 & 57.10 & 66.63\\  
        & F1↑  & \textbf{69.6}  & 91.03 & 84.49 & 72.78 & \textbf{84.47} &  91.30 & 77.86 & 81.65\\  
        & ED↓  & 500   & 779   & 6632  & 118    & 350   &  136 & 1538 & 1436.14\\  
        \hline

       %   \multirow{3}{*}{Singletask + LLM (50\% traindata size)}  
       % & ACC↑ & 48.54 & 80.64 & 76.77 & 52.29 & 67.02 &  81.44 & 58.98 & 66.81 \\  
      %  & F1↑  & 67.86 & 89.81 & 82.61 & 67.43 & 82.84 &  90.84 & 78.76 & 80.31\\  
      %  & ED↓  & 518   & 873  & 7037  & 101   & 367   &  \textbf{127} & 1441 & 1480.57\\  
     %   \hline
        
         \multirow{3}{*}{M + LLM (0.5)}  
        & ACC↑ & \textbf{49.51} & 81.64 & 78.41 & 52.29 & 67.02 & 80.84 & 56.89 & 66.66 \\  
        & F1↑  & 67.44  & 91.87 & 84.91 & 70.84 & 82.84 & 90.45 & 76.97 & 80.76 \\  
        & ED↓  & 529   & \textbf{687}   & 6483  & 117   & 367   & 164   & 1557 & 1414.86 \\  
        \hline
        
        %  \multirow{3}{*}{Singletask + LLM (75\% traindata size)}  
        % & ACC↑ & 48.57& 80.59 & 78.27 & 55.05 & \textbf{72.47} &  80.24 & \textbf{59.87} & \textbf{67.44}\\  
       % & F1↑  & 67.71  & 86.02 & 83.98 & 70.72 & 84.40 &  90.01 & 78.97 & 80.54 \\  
      %  & ED↓  & 536   & 863  & 6635  & 99   & \textbf{287}  &  147 & \textbf{1432} & 1428.43\\  
       % \hline
        
          \multirow{3}{*}{M+ LLM (0.75)}  
        & ACC↑ & 48.82 & 81.32 & \textbf{79.5}  & \textbf{56.88} & 68.71 & \textbf{81.24} & \textbf{58.29} & \textbf{67.82} \\  
        & F1↑  & 67.69 & 91.51 & \textbf{85.65} & \textbf{74.32} & 84.18 & 91.34 & \textbf{79.05} & \textbf{81.96} \\  
        & ED↓  & \textbf{491}   & 723   & \textbf{6502}  & \textbf{96}    & \textbf{333}   & \textbf{127}   & \textbf{1507} &  \textbf{1397}\\  
        \hline
    \end{tabular}
    \end{adjustbox}
    \caption{Comparison of segmentation models across languages. Each model includes three sub-rows for ACC, F1, and ED, with the last column showing average metrics. M denotes multitask learning, with synthetic data added at 25\%, 50\%, and 75\% of training size.}
    \label{tab:LLM_results}
\end{table}

LLM-generated synthetic data can be highly beneficial in addressing the data scarcity problem for canonical segmentation. By providing diverse and linguistically plausible training examples, LLMs help compensate for the lack of annotated data while preserving the structural integrity of morphological patterns. The improvements observed in both accuracy and consistency demonstrate the value of incorporating LLMs into segmentation models, particularly for languages with limited annotated resources.  We have constrained our presented experiments to the multitask setting, but an ablation study on the single-task objective (Appendix \ref{single-ablation}) demonstrates similar trends.

%For further evidence, we present the results of our ablation experiments for the single-task setting in the Appendix \ref{single-ablation}. The findings generally align with those of the multi-task approach.

\section{Conclusions}
In this work, we have demonstrated that low-resource canonical morpheme segmentation is improved through the use of multitask learning and synthetic data. Using glossing as an auxiliary task and LLMs to strengthen the training signal, we provide a new benchmark for canonical morpheme segmentation in low-resource languages, aiding in the development of effective computational tools for linguistic documentation and preservation. 
%Our findings highlight that leveraging existing linguistic annotations and synthetic data mitigates data scarcity while maintaining morphological consistency. 
Future research should refine data augmentation techniques, explore active learning strategies, and investigate multilingual training frameworks to improve cross-linguistic generalization, while also working with documentary linguists to evaluate the value of automation in the field. %This work advances the development of effective computational tools for linguistic documentation and preservation.

\section{Limitations}
Despite the improvements demonstrated in our experiments, our approach has several limitations that should be addressed in future research. One key limitation is our reliance on synthetic data generated by large language models (LLMs). While we observe performance gains when augmenting training with synthetic examples, the quality and linguistic validity of these examples remain uncertain. LLMs may introduce hallucinations, generating segmentation patterns that do not fully align with the true morphological structure of the target language. Since our study does not include a detailed qualitative error analysis, it is difficult to determine whether the improvements stem from genuinely better morphological generalization or simply from increased exposure to frequent patterns. A more thorough investigation of the impact of synthetic data on segmentation quality, particularly in low-resource settings, is necessary. 

One potential risk of LLM-generated synthetic data lies in the misuse of these data for deceptive or unethical purposes. Since we propose using LLMs to generate structured linguistic data, this technique could be exploited to fabricate linguistic evidence in historical or sociolinguistic studies. In particular, if synthetic morphological data is presented as authentic, it could be used to falsely attribute linguistic features to certain languages or communities, potentially leading to misrepresentation or erasure of genuine linguistic diversity. 

A second limitation is that because our synthetic data generation process relies on patterns observed in the training set, it is inherently limited to existing vocabulary. The LLM-generated data cannot create new stems or morphological categories that have not appeared in the training data, restricting its ability to model truly novel linguistic forms. This limitation means that the model may still struggle with out-of-vocabulary (OOV) words or rare morphological constructions that were not adequately represented in the original dataset. Future research could explore alternative methods, such as leveraging morphological rule induction or few-shot learning with human-in-the-loop guidance, to generate more diverse and linguistically valid synthetic data that extends beyond what has been seen in the training set.

\section{Ethical Concerns}
As with any work involving language data, but particularly data from underserved and historically marginalized communities, steps should be taken that language corpora are collected and stewarded with respect and the support of the communities. These data represent the linguistic and cultural heritage of communities of people, and we thank the people of these communities for allowing us to work with their languages.

\bibliography{custom,anthology}

\begin{thebibliography}{13}
\expandafter\ifx\csname natexlab\endcsname\relax\def\natexlab#1{#1}\fi

\bibitem[{Caruana(1997)}]{caruana1997multitask}
Rich Caruana. 1997.
\newblock Multitask learning.
\newblock \emph{Machine learning}, 28:41--75.

\bibitem[{Ginn et~al.(2023)Ginn, Moeller, Palmer, Stacey, Nicolai, Hulden, and Silfverberg}]{ginn2023findings}
Michael Ginn, Sarah Moeller, Alexis Palmer, Anna Stacey, Garrett Nicolai, Mans Hulden, and Miikka Silfverberg. 2023.
\newblock Findings of the {SIGMORPHON} 2023 shared task on interlinear glossing.
\newblock In \emph{Proceedings of the 20th SIGMORPHON workshop on Computational Research in Phonetics, Phonology, and Morphology}, pages 186--201.

\bibitem[{Johnson et~al.(2017)Johnson, Schuster, Le, Krikun, Wu, Chen, Thorat, Vi{\'e}gas, Wattenberg, Corrado, Hughes, and Dean}]{johnson-etal-2017-googles}
Melvin Johnson, Mike Schuster, Quoc~V. Le, Maxim Krikun, Yonghui Wu, Zhifeng Chen, Nikhil Thorat, Fernanda Vi{\'e}gas, Martin Wattenberg, Greg Corrado, Macduff Hughes, and Jeffrey Dean. 2017.
\newblock \href {https://doi.org/10.1162/tacl_a_00065} {{G}oogle`s multilingual neural machine translation system: Enabling zero-shot translation}.
\newblock \emph{Transactions of the Association for Computational Linguistics}, 5:339--351.

\bibitem[{Kann et~al.(2016)Kann, Cotterell, and Sch{\"u}tze}]{kann-etal-2016-neural}
Katharina Kann, Ryan Cotterell, and Hinrich Sch{\"u}tze. 2016.
\newblock \href {https://doi.org/10.18653/v1/D16-1097} {Neural morphological analysis: Encoding-decoding canonical segments}.
\newblock In \emph{Proceedings of the 2016 Conference on Empirical Methods in Natural Language Processing}, pages 961--967, Austin, Texas. Association for Computational Linguistics.

\bibitem[{Lin et~al.(2018)Lin, Yang, Stoyanov, and Ji}]{lin-etal-2018-multi-lingual}
Ying Lin, Shengqi Yang, Veselin Stoyanov, and Heng Ji. 2018.
\newblock \href {https://doi.org/10.18653/v1/P18-1074} {A multi-lingual multi-task architecture for low-resource sequence labeling}.
\newblock In \emph{Proceedings of the 56th Annual Meeting of the Association for Computational Linguistics (Volume 1: Long Papers)}, pages 799--809, Melbourne, Australia. Association for Computational Linguistics.

\bibitem[{Ott et~al.(2019)Ott, Edunov, Baevski, Fan, Gross, Ng, Grangier, and Auli}]{ott-etal-2019-fairseq}
Myle Ott, Sergey Edunov, Alexei Baevski, Angela Fan, Sam Gross, Nathan Ng, David Grangier, and Michael Auli. 2019.
\newblock \href {https://doi.org/10.18653/v1/N19-4009} {fairseq: A fast, extensible toolkit for sequence modeling}.
\newblock In \emph{Proceedings of the 2019 Conference of the North {A}merican Chapter of the Association for Computational Linguistics (Demonstrations)}, pages 48--53, Minneapolis, Minnesota. Association for Computational Linguistics.

\bibitem[{Rice et~al.(2024)Rice, Marashian, Gessler, Palmer, and von~der Wense}]{rice-etal-2024-tams}
Enora Rice, Ali Marashian, Luke Gessler, Alexis Palmer, and Katharina von~der Wense. 2024.
\newblock \href {https://doi.org/10.18653/v1/2024.acl-long.366} {{TAMS}: Translation-assisted morphological segmentation}.
\newblock In \emph{Proceedings of the 62nd Annual Meeting of the Association for Computational Linguistics (Volume 1: Long Papers)}, pages 6752--6765, Bangkok, Thailand. Association for Computational Linguistics.

\bibitem[{Ruzsics and Samard{\v{z}}i{\'c}(2017)}]{ruzsics-samardzic-2017-neural}
Tatyana Ruzsics and Tanja Samard{\v{z}}i{\'c}. 2017.
\newblock \href {https://doi.org/10.18653/v1/K17-1020} {Neural sequence-to-sequence learning of internal word structure}.
\newblock In \emph{Proceedings of the 21st Conference on Computational Natural Language Learning ({C}o{NLL} 2017)}, pages 184--194, Vancouver, Canada. Association for Computational Linguistics.

\bibitem[{Vaswani et~al.(2017)Vaswani, Shazeer, Parmar, Uszkoreit, Jones, Gomez, Kaiser, and Polosukhin}]{vaswani2017attention}
Ashish Vaswani, Noam Shazeer, Niki Parmar, Jakob Uszkoreit, Llion Jones, Aidan~N Gomez, {\L}ukasz Kaiser, and Illia Polosukhin. 2017.
\newblock Attention is all you need.
\newblock \emph{Advances in neural information processing systems}, 30.

\bibitem[{Wang et~al.(2019)Wang, Fam, Bao, Lepage, and Gao}]{wang2019neural}
Weihua Wang, Rashel Fam, Feilong Bao, Yves Lepage, and Guanglai Gao. 2019.
\newblock Neural morphological segmentation model for mongolian.
\newblock In \emph{2019 International Joint Conference on Neural Networks (IJCNN)}, pages 1--7. IEEE.

\bibitem[{Wiseman and Rush(2016)}]{wiseman2016sequence}
Sam Wiseman and Alexander~M Rush. 2016.
\newblock Sequence-to-sequence learning as beam-search optimization.
\newblock In \emph{Proceedings of the 2016 Conference on Empirical Methods in Natural Language Processing}, pages 1296--1306.

\bibitem[{Wu et~al.(2021)Wu, Cotterell, and Hulden}]{wu-etal-2021-applying}
Shijie Wu, Ryan Cotterell, and Mans Hulden. 2021.
\newblock \href {https://doi.org/10.18653/v1/2021.eacl-main.163} {Applying the transformer to character-level transduction}.
\newblock In \emph{Proceedings of the 16th Conference of the European Chapter of the Association for Computational Linguistics: Main Volume}, pages 1901--1907, Online. Association for Computational Linguistics.

\bibitem[{Zhou et~al.(2019)Zhou, Zeng, Zhou, Anastasopoulos, and Neubig}]{zhou-etal-2019-improving}
Shuyan Zhou, Xiangkai Zeng, Yingqi Zhou, Antonios Anastasopoulos, and Graham Neubig. 2019.
\newblock \href {https://doi.org/10.18653/v1/W19-5368} {Improving robustness of neural machine translation with multi-task learning}.
\newblock In \emph{Proceedings of the Fourth Conference on Machine Translation (Volume 2: Shared Task Papers, Day 1)}, pages 565--571, Florence, Italy. Association for Computational Linguistics.

\end{thebibliography}


\appendix
\section{Appendix}
\subsection{Model Hyperparameters}
\label{model-appendix}
We train our models with 4 layers in each encoder and 2 or 4 layers in each decoder, each containing 4 attention heads. The embedding size is 256 and the hidden layer size is 1024. These
hyper-parameter settings roughly correspond to the values used by \cite{wu-etal-2021-applying} for character-level tasks. We use the Adam optimizer with an initial learning rate of 0.001,with both dropout and attention dropout set to 0.1, and batch size 400. We train the model for 150-300 epochs and the prediction is performed with the best checkpoint model, according to the development accuracy, using a beam of width 5. 

\subsection{Effect of \(\lambda\) Weighting on Multitask Performance}
\label{weighting}

Table~\ref{tab:weighting} illustrates the impact of adjusting the segmentation-glossing weight (\(\lambda\)) on Lezgi model performance. As \(\lambda\) increases, placing greater emphasis on segmentation loss, both accuracy and morpheme-level F1-score improve consistently.

These results suggest that balancing segmentation and glossing loss is crucial for multitask learning effectiveness. While high values of \(\lambda\) are generally beneficial, completely discarding glossing supervision could lead to the loss of valuable linguistic information. Thus, fine-tuning \(\lambda\) is essential to achieve the best trade-off between segmentation precision and linguistic generalization.

\begin{table}[h]
    \centering
    \begin{adjustbox}{width=\columnwidth,center}
        \begin{tabular}{lcc}
            \hline
            Model & Accuracy (\%) & F1-score (\%) \\
            \hline
            Single-task Baseline & 44.66 & 60.75 \\
            Multitask (\(\lambda=0.5\)) & 40.78 & 62.12 \\
            Multitask (\(\lambda=0.6\)) & 42.20 & 63.31 \\
            Multitask (\(\lambda=0.7\)) & 43.23 & 65.59 \\
            Multitask (\(\lambda=0.8\)) & 46.23 & 66.59 \\
            Multitask (\(\lambda=0.9\)) & 48.54 & 68.84 \\
            Multitask (\(\lambda=1\)) & 48.04 & 68.12 \\
            \hline
        \end{tabular}
    \end{adjustbox}
    \caption{Impact of \(\lambda\) weighting on Lezgi model performance.}
    \label{tab:weighting}
\end{table}

\subsection{Learning Curves among All Languages}
\label{learning-curve-all}
\begin{figure*}
\includegraphics[width=\textwidth]{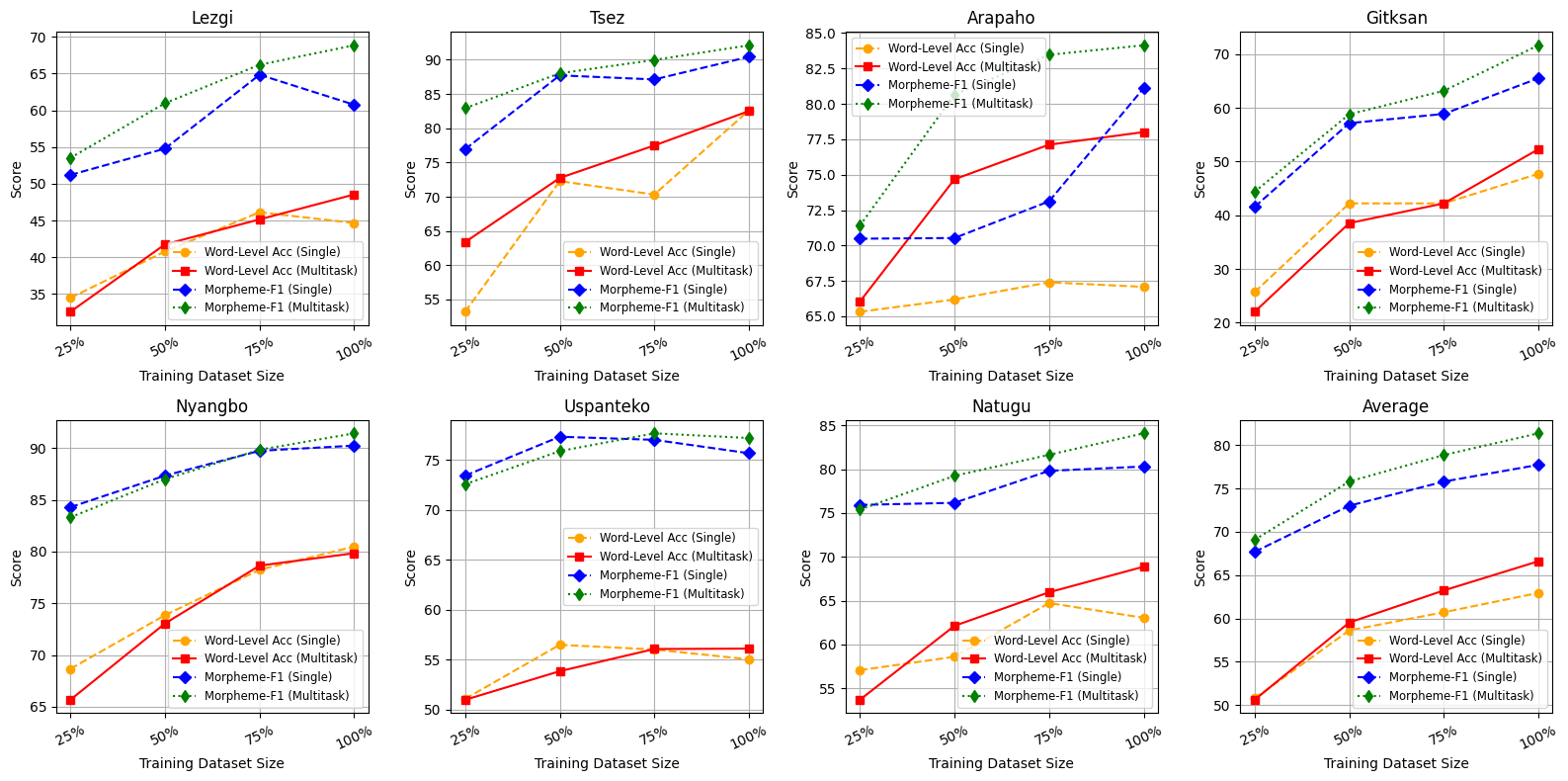}
\caption{The learning curves for the F1 (top)
and Accuracy (bottom) metrics among all languages.}
\label{all-lc}
\end{figure*}

Figure~\ref{all-lc} presents learning curves across different training dataset sizes (25\%, 50\%, 75\%, and 100\%). Each subplot corresponds to a different language, with the final panel showing the average trends across all languages.

Across all languages, the multitask model (solid lines) consistently outperforms the single-task model (dashed lines), particularly at lower training data sizes. This trend is most pronounced in Lezgi, Gitksan, and Arapaho, where multitask learning significantly boosts both word-level accuracy (red squares vs. orange circles) and morpheme F1-score (green diamonds vs. blue diamonds). 

For languages like Nyangbo and Tsez, the difference between single-task and multitask learning diminishes as dataset size increases. Additionally, while morpheme F1-score improves steadily with more training data, word-level accuracy plateaus earlier in some languages (e.g., Uspanteko, Nat\"ugu), suggesting that segmentation benefits more from additional data than word-level reconstruction does.

\subsection{LLM Prompt}
\label{LLM-prompt-example}
You are a linguistics expert of Nat\"ugu. Your job is to generate new words based on the examples you learned. You are given this stem "pr", its meaning is "go". Here are several word examples of this stems: 

Example 1:

surface form: prtrp, canonical segmentation: pr-tr-mq, gloss: go-GDIR.IN-PDIR.HITHER

......

You are also given a list of grammatical morphemes and their corresponding gloss: 

Grammatical gloss "3AUG", its morpheme is "nz"

Grammatical gloss "COS", its morpheme is "pe". 

......

Can you generate 3 new words using the stem and randomly use 2-5 grammatical morphemes. You need to return the result in the same format as the examples (word, canonical segmentation, and gloss). Please note that canonical segmentation will have character change. 

\subsection{Single-Task Ablation Results}
\label{single-ablation}

Table \ref{tab:LLM_results} presents an ablation study evaluating the impact of LLM-generated synthetic data on both single-task and multitask models for canonical segmentation. Across all languages, adding synthetic data consistently improves segmentation performance, particularly at the morpheme level (F1-score). Notably, for single-task models, synthetic data provides incremental improvements, but these gains are more pronounced in the multitask setting, where segmentation and glossing are jointly learned.

When comparing S+LLM (0.5) vs. M+LLM (0.5) we observe that multitask learning consistently outperforms single-task learning across all metrics. The average F1-score for the multitask model (80.76\%) is higher than the single-task model (80.02\%), and the edit distance (ED) is also reduced more effectively (1414.86 vs. 1480.57). This suggests that multitask learning better integrates synthetic data, leveraging glossing as an auxiliary task to reduce segmentation errors and improve consistency.

Interestingly, in lower-resource languages like Gitksan, LLM augmentation provides the largest gains, particularly at higher proportions (75\%), reinforcing that synthetic data is most beneficial in extreme data-scarce conditions. However, for languages with richer training data like Tsez and Nyangbo, improvements plateau.

\begin{table}[h]
\begin{adjustbox}{width=\columnwidth,center}
    \centering
    \begin{tabular}{|l|l|c|c|c|c|c|c|c|c|}
        \hline
        \textbf{Model} & \textbf{Metric} & \textbf{lez} & \textbf{ddo} & \textbf{arp} & \textbf{git} & \textbf{ntu} & \textbf{nyb} & \textbf{usp} & \textbf{ave} \\
        \hline
      \multirow{3}{*}{Baseline (S)}  
        & ACC↑ & 44.66 & \textbf{82.6} & 67.08 & 47.71 & 63.04 & 80.48 & 55.05 & 62.95 \\  
       & F1↑  & 60.75 & 90.44 & 81.11 & 65.5  & 80.3  & 90.24 & 75.66 & 77.71 \\  
        & ED↓  & 568   & \textbf{652}   & 10495 & 117   & 458   & 154   & 1799 & 2034.71 \\  
        \hline
     %   \multirow{3}{*}{TAMS}  
    %    & ACC↑ & 46.84 & 80.78 & 67.72 & -- & -- & -- & -- & -- \\  
    %    & F1↑  & 62.48 & 89.52 & 81.62 & -- & -- & -- & -- & -- \\  
    %    & ED↓  & 532   & 701   & 9899  & -- & -- & -- & -- & -- \\  
     %   \hline
    %    \multirow{3}{*}{TAMS-CLS}  
    %    & ACC↑ & 47.09 & 81.96 & 67.4  & -- & -- & -- & -- & --\\  
    %    & F1↑  & 62.48 & 90.08 & 81.48 & -- & -- & -- & -- & --\\  
    %    & ED↓  & 537   & 648   & 9970  & -- & -- & -- & -- & --\\  
     %   \hline
        \multirow{3}{*}{M}  
        & ACC↑ & 48.54 & 82.51 & 78.01 & 52.29 & 68.87 & 79.84 & 56.12 & 66.59 \\  
        & F1↑  & 68.84 & \textbf{92.12} & 84.14 & 71.64 & 84.09 & \textbf{91.43} & 77.18 & 81.35 \\  
        & ED↓  & 519   & 698   & 6543  & 112   & 373   & 149   & 1623 & 1431 \\  
        \hline 
        %\multirow{3}{*}{Only CRF}  
       % & ACC↑ & 49.76 & 80.0  & 71.1  & 53.21 & 70.25 & 80.84 & 57.61 & 66.11 \\  
        %& F1↑  & 67.04 & 90.96 & 76.91 & 68.8  & 83.48 & 90.77 & 78.47 & 79.20 \\  
       % & ED↓  & 521   & 783   & 9401  & 96    & 328   & 142   & 1492 & 1623.29 \\  
        %\hline
      %  \multirow{3}{*}{multi-task (0.8, 0.2) + CRF}  
      %  & ACC↑ & 47.33 & 80.32 & 77.0  & 50.46 & 67.33 & 78.24 & 56.2 & 65.99 \\  
       % & F1↑  & 67.29 & 91.34 & 83.68 & 68.21 & 82.64 & 90.17 & 78.06 & 80.48 \\  
       % & ED↓  & 556   & 772   & 6844  & 126   & 398   & 163   & 1604 & 1623.29 \\  
      %  \hline
         \multirow{3}{*}{S+ LLM (0.25)}  
       & ACC↑ & 48.79 & 80.28 & 78.17 & 53.96 & 66.10&  80.04 & 56.50 & 66.26\\  
        & F1↑  & 68.17 & 90.82& 83.96 & 70.95 & 80.55 &  90.73 & 76.59 & 80.25\\  
        & ED↓  &475   & 852  & 6534 & \textbf{92}    & 357   &  137 & 1544& 1427.29\\  
        \hline
        
        \multirow{3}{*}{M+ LLM (0.25)}  
        & ACC↑ & 49.27 & 80.41 & 78.14  & 52.29 & 69.02 &  80.21 & 57.10 & 66.63\\  
        & F1↑  & \textbf{69.6}  & 91.03 & 84.49 & 72.78 & \textbf{84.47} &  91.30 & 77.86 & 81.65\\  
        & ED↓  & 500   & 779   & 6632  & 118    & 350   &  136 & 1538 & 1436.14\\  
        \hline

         \multirow{3}{*}{S+ LLM (0.5)}  
        & ACC↑ & 48.54 & 80.64 & 76.77 & 52.29 & 67.02 &  81.44 & 58.98 & 66.52 \\  
       & F1↑  & 67.86 & 89.81 & 82.61 & 67.43 & 82.84 &  90.84 & 78.76 & 80.02\\  
       & ED↓  & 518   & 873  & 7037  & 101   & 367   &  \textbf{127} & 1441 & 1480.57\\  
       \hline
        
         \multirow{3}{*}{M + LLM (0.5)}  
        & ACC↑ & \textbf{49.51} & 81.64 & 78.41 & 52.29 & 67.02 & 80.84 & 56.89 & 66.66 \\  
        & F1↑  & 67.44  & 91.87 & 84.91 & 70.84 & 82.84 & 90.45 & 76.97 & 80.76 \\  
        & ED↓  & 529   & 687   & 6483  & 117   & 367   & 164   & 1557 & 1414.86 \\  
        \hline
        
         \multirow{3}{*}{S + LLM (0.75)}  
         & ACC↑ & 48.57& 80.59 & 78.27 & 55.05 & \textbf{72.47} &  80.24 & \textbf{59.87} & \textbf{67.87}\\  
        & F1↑  & 67.71  & 86.02 & 83.98 & 70.72 & 84.40 &  90.01 & 78.97 & 80.26 \\  
        & ED↓  & 536   & 863  & 6635  & 99   & \textbf{287}  &  147 & \textbf{1432} & 1428.43\\  
        \hline
        
          \multirow{3}{*}{M+ LLM (0.75)}  
        & ACC↑ & 48.82 & 81.32 & \textbf{79.5}  & \textbf{56.88} & 68.71 & \textbf{81.24} & 58.29 & 67.82 \\  
        & F1↑  & 67.69 & 91.51 & \textbf{85.65} & \textbf{74.32} & 84.18 & 91.34 & \textbf{79.05} & \textbf{81.96} \\  
        & ED↓  & \textbf{491}   & 723   & \textbf{6502}  & 96    & 333   & \textbf{127}   & 1507 &  \textbf{1397}\\  
        \hline
    \end{tabular}
    \end{adjustbox}
    \caption{Comparison of segmentation models across multiple languages. Each model has three sub-rows representing Word-Level Accuracy (ACC), Morpheme F1-Score (F1), and Edit Distance (ED). The last column provides the average of each metric across languages. M denotes multitask learning, and S denotes single-task learning, with synthetic data added at 25\%, 50\%, and 75\% of training size.}
    \label{tab:LLM_results}
\end{table}

\end{document}